\documentclass[10pt,twocolumn,letterpaper]{article}

\usepackage{iccv}
\usepackage{times}
\usepackage{epsfig}
\usepackage{graphicx}
\usepackage{amsmath}
\usepackage{amssymb}

\usepackage{multirow}
\usepackage{makecell}
\usepackage{helvet}
\usepackage{courier}
\usepackage{graphicx}
\usepackage{bm}
\usepackage{color}
\usepackage{epstopdf}
\usepackage{caption}
\usepackage{subcaption}
\usepackage{enumitem}
\usepackage{calc}
\usepackage{multirow}
\usepackage{xspace}
\usepackage{booktabs}
\usepackage{mathrsfs}
\usepackage{array}
\usepackage{gensymb}
\usepackage{bbm}
\usepackage{dsfont}
\usepackage{bm}

\usepackage{multirow}
\usepackage{subcaption}

\usepackage{amsmath,lipsum}
\usepackage{cuted}
\usepackage{comment}
\usepackage[font=small,labelfont=bf]{caption}

\newcommand{\R}[1]{\textcolor{red}{#1}}

\newcommand{\B}[1]{\textcolor{blue}{#1}}

\usepackage[pagebackref=true,breaklinks=true,letterpaper=true,colorlinks,bookmarks=false]{hyperref}

\iccvfinalcopy 

\def\iccvPaperID{8478} 

\ificcvfinal\fi

\begin{document}

\title{AutoShape: Real-Time Shape-Aware Monocular 3D Object Detection}

\author{Zongdai Liu,\quad Dingfu Zhou \thanks{Corresponding author}, \quad Feixiang Lu,\quad Jin Fang and Liangjun Zhang\\
\small Robotics and Autonomous Driving Laboratory, Baidu Research, \\ 
\small National Engineering Laboratory of Deep Learning Technology and Application, China \quad \\
{\tt\small \{liuzongdai, zhoudingfu, lufeixiang, fangjin, liangjunzhang\}@baidu.com}
}

\maketitle
\ificcvfinal\thispagestyle{empty}\fi

\begin{abstract}

Existing deep learning-based approaches for monocular 3D object detection in autonomous driving often model the object as a rotated 3D cuboid while the object's geometric shape has been ignored. In this work, we propose an approach for incorporating the shape-aware 2D/3D constraints into the 3D detection framework. Specifically, we employ the deep neural network to learn distinguished 2D keypoints in the 2D image domain and regress their corresponding 3D coordinates in the local 3D object coordinate first. Then the 2D/3D geometric constraints are built by these correspondences for each object to boost the detection performance. For generating the ground truth of 2D/3D keypoints, an automatic model-fitting approach has been proposed by fitting the deformed 3D object model and the object mask in the 2D image. The proposed framework has been verified on the public KITTI dataset and the experimental results demonstrate that by using additional geometrical constraints the detection performance has been significantly improved as compared to the baseline method. More importantly, the proposed framework achieves state-of-the-art performance with real time. Data and code will be available at \url{https://github.com/zongdai/AutoShape}

\end{abstract}

\section{Introduction}\label{sec:intro}

Perceiving 3D shapes and poses of surrounding obstacles is an essential task in autonomous driving (AD) perception systems. The accuracy and speed performance of 3D objection detection is important for the following motion planning and control modules in AD. Many 3D object detectors \cite{zhou2018voxelnet, lang2019pointpillars} have been proposed, mainly for depth sensors such as LiDAR \cite{shi2019pointrcnn, zhou2020joint} or stereo cameras \cite{zhou2014modeling, li2019stereo}, which can provide the distance information of the environments directly. However, LiDAR sensors are expensive and stereo rigs suffer from on-line calibration issues. 
Therefore, monocular camera based 3D object detection becomes a promising direction.


The main challenge for monocular-based approaches is to obtain accurate depth information. In general, depth estimation from a single image without any prior information is a challenging problem and recent many deep learning-based approaches achieve good results \cite{godard2019digging}. With the estimated depth map, pseudo LiDAR point cloud can be reconstructed via pre-calibrated intrinsic camera parameters and 3D detectors designed for LiDAR point cloud can be applied directly on pseudo LiDAR point cloud \cite{qian2020end} \cite{vianney2019refinedmpl}. Furthermore, \cite{qian2020end} integrates the depth estimation and 3D object detection network together following an end-to-end manner. However, heavy computation burden is one main bottleneck of such two-stage approaches. 

\begin{figure}
\centering
\includegraphics[width=0.98\linewidth]{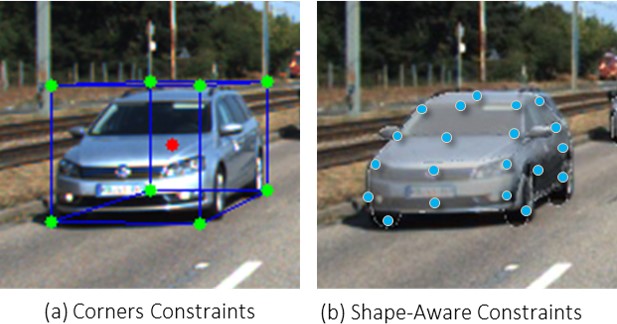}
\caption{(a): 3D Bbox corners and center are commonly used for monocular 3D object detection. However, the rich structure information from 3D vehicle shapes and their projection on 2D images are not employed. (b) shows our shape-aware constraints constructed from an aligned 3D model. Such 2D-3D keypoints carry more semantic and geometric information and enable to construct stronger geometric constraints for monocular 3D detection.}
\label{Fig:Teaser}
\end{figure}


To improve the efficiency, many direct regression-based approaches have been proposed (\eg, SMOKE~\cite{liu2020smoke}, RTM3D~\cite{li2020rtm3d}) and achieved promising results. By representing the object as one center point, the object detection task is formulated as keypoints detection and its corresponding attributes (\eg, size, offsets, orientation, depth, \etc) regression. With this compact representation, the computation speed of this kind of approach can reach 20$\sim$30 fps (frame per second). However, the drawback is also obvious. One center point \cite{zhou2019objects, liu2020smoke} representation ignores the detailed shape of the object and results in location ambiguity if its projected center point is on another object's surface due to occlusion \cite{zhou2020iafa}. 
To alleviate this ambiguity, other geometrical constraints have been used to improve the performance. RTM3D~\cite{li2020rtm3d} adds 8 more keypoints as additional constraints which are defined as the projected 2d location of the 3D bounding box's corners. However, these keypoints have non-real context meanings and their 2D locations vary differently with the changing of the camera view-point, even the object's orientation. As shown in the left of Fig.~\ref{Fig:Teaser}, some keypoints are on the ground and some are on the sky or trees. This makes the keypoints detection network extremely difficult to distinguish the keypoints or other image pixels.

In this paper, we propose a novel approach to learn the meaningful keypoints on the object surface and then use them as additional geometrical constraints for 3D object detection. Specifically, we design an automatic deformable model-fitting pipeline first to generate the 2D/3D correspondences for each object. Then, the center point plus several distinguished keypoints are learned from the deep neural network. Based on these keypoints and other regressed objects' attributes (\eg, orientation angle, object dimension \etc), the object's 3D bounding box can be solved with linear equations. The proposed framework can be trained in an end-to-end manner. Our contributions include:

\begin{enumerate}[leftmargin=*]
\item We propose a shape-aware 3d object detection framework, which employs keypoints geometry constraints for 2D/3D regression to boost the detection performance.
\item We present a method for automatically fitting the 3D shape to the visual observations and then generating ground-truth annotations of 2D/3D keypoints pairs for the training network. Our source code and dataset will be made public for the community.
\item The effectiveness of our approach has been verified on the public KITTI dataset and achieved SOTA performance. 
More importantly, the proposed framework achieves real-time ($25$ fps), which can be integrated into the AD perception module. 
\end{enumerate}

\section{Related Work} \label{sec:related_work}

\subsection{Monocular-based 3D Detection}%

Image-based 3D object detection becomes popular due to the cheap price of the camera sensors. Stereo-based approaches usually suffer from calibration issues between two camera rigs. Therefore, many 3D object detection approaches have proposed to use a single image frame. Generally, these approaches can be categorized into three types: depth-map-based, direct regression-based, and CAD model-based methods. 

Depth-map-based methods \cite{wang2019task} usually need to estimate the depth map first. In \cite{weng2019monocular} and \cite{vianney2019refinedmpl}, the estimated depth map is transformed into point clouds, and then point-cloud-based 3D object detectors are employed for achieving the detection results. Rather than transforming the depth map into point clouds, many approaches propose using the depth estimation map directly in the framework to enhance the 3D object detection. In M3D-RPN \cite{brazil2019m3d} and \cite{ding2020learning}, the pre-estimated depth map has been used to guide the 2D convolution, which is called as ``Depth-Aware Convolution''. Direct regression-based methods are proposed to estimate the objects' 3D information via image domain directly, such as \cite{li2019gs3d,liu2020smoke,zhou2019objects,zhou2020iafa}. Direct-based methods are much more efficient than depth-map-based methods because the depth-map computation procedure is not necessary.

In order to well benefit the prior knowledge, the shape information has been integrated into the CAD-based approaches. Deep MANTA \cite{chabot2017deep} and ApolloCar3D \cite{song2019apollocar3d} are two keypoints based methods, in which the 3D keypoints are pre-defined on the CAD model and their corresponding 2D points on the image plane are computed by the deep neural network. Then the 3D pose can be solved with a standard 2D/3D pose solver \cite{lepetit2009epnp} with these 2D/3D correspondences. Besides keypoints-based methods, dense-matching-based approaches are proposed in \cite{kundu20183d, ku2019monocular, manhardt2019roi}. In \cite{kundu20183d}, Rendering-and-Compare loss is designed for optimizing the 3d pose estimation. While in \cite{ku2019monocular} and \cite{manhardt2019roi}, the 3D pose estimation and reconstruction of each object are generated simultaneously with the deep neural network.    

\subsection{Data Labeling for 3D Object Detection}
For easy representation, objects are usually described as 3D cuboids in deep learning frameworks while the shape information has been totally ignored. Manually label the object shape via only the image observation is extremely difficult and the annotation quality also can not be guaranteed. Many CAD model guided annotation approaches have been proposed to obtain the dense shape annotations. In \cite{menze2015object}, both the stereo image and the sparse LiDAR point cloud has been employed for generating the dense scene flow for both foreground and background pixels. For dynamic objects, 16 vehicle models are chosen as basic templates and then the dense annotation is achieved by finding an optimal 3D similarity transformation (e.g., the pose and scale of the 3D model) with three types of observations such as LiDAR points, dense disparity computed by SGM \cite{hirschmuller2005accurate} and labeled 2D/3D correspondences. 

In \cite{song2019apollocar3d}, 66 keypoints are defined on the 3D CAD models and annotators label their corresponding 2D keypoints on the image. Based on the 2D/3D correspondences, the object poses can be obtained via a PnP solver. In \cite{zakharov2020autolabeling}, the authors apply a differentiable shape renderer to signed distance fields (SDF), leveraged together with normalized object coordinate spaces (NOCS) to automatically generate the dense 3D shape without the 3D bounding boxes annotation. Although the whole process is labor-free, the annotation quality is far-from the ground truth. Different from \cite{zakharov2020autolabeling}, we use the ground truth 3D bounding boxes as strong guidance for our 3D shape annotation generation process.





\section{Problem Definition}\label{basic_knowledge}
Before the introduction of our proposed approach, a general description of image-based 3D object detection problem is introduced first.


\subsection{Pose Estimation}
Given an image, the task of \textbf{pose estimation} is to estimate the orientation and translation of objects in 3D. Specifically, 6D pose is represented by a rigid transformation $(\mathbf{R}, \mathbf{T})$ from the \textit{object coordinate system} to the \textit{camera coordinate system}, where $\mathbf{R}$ represents the 3D rotation and $\mathbf{T}$ represents the 3D translation. 

Assuming a 3D object point $\mathbf{P}_{o}$ $(x_o, y_o, z_o)$ in the object coordinate system, transformed 3d point $\mathbf{P}_{c}$ $(x_c, y_c, z_c)$  in camera coordinate can be obtained as  
\begin{equation}
 [x_c, y_c, z_c]^{T} =  \mathbf{R} [x_o, y_o, z_o]^{T} + \mathbf{T},
 \label{Eq:o2c}
\end{equation}
where $\mathbf{R}$ is rotation matrix, $\mathbf{T}$ is translation vector. Given the camera intrinsic matrix 
\small 
$\mathbf{K}= \left [ \setlength{\arraycolsep}{1.1pt}        \begin{array}{ccc}
        f_x & 0  & c_x  \\
        0 & f_y &  c_y  \\
        0 & 0 &  1  \\
    \end{array} \right ]$
\normalsize
, the projected image point $\mathbf{p}$ $(u, v)$ can be obtained as  
\begin{equation}
 s[u, v, 1]^{T} =  \mathbf{K} [ x_c, y_c, z_c]^{T}.
 \label{Eq:c2imag}
\end{equation}


Based on Eq. \ref{Eq:o2c} and Eq. \ref{Eq:c2imag}, the object pose $\mathbf{R}$ and $\mathbf{T}$ can be theoretically recovered with the geometric constraints between 3D points $\mathbf{P}_{o}$ on the object and the projected 2D image points $\mathbf{p}$. 

\subsection{Learning-based 3D Object Detection}
In the era of deep learning, many approaches have been proposed to detect objects and directly regress their poses using neural networks, while geometric 2D/3D constraints have been ignored in the formulation. Image-based 3D object detection is a typical task, which aims at estimating the location, orientation of an object in the camera coordinate. Usually, an object is represented as a rotated 3D BBox as 
\begin{equation}
   \textbf{r}=(r_x, r_y, r_z); \textbf{t}=(t_x, t_y, t_z); \textbf{d} = (l, w, h),
\end{equation}
in which $\textbf{r}$, $\textbf{t}$ represent the object's orientation, location in the camera coordinate and $\textbf{d}$ is the dimension of the object. With the super expression ability of neural networks, all these parameters are regressed directly without imposing addition constraints. Indeed, both 3D object detection and pose estimation are essentially the same problem and ($\textbf{r}$, $\textbf{t}$) can be easily transformed from ($\textbf{R}$, $\textbf{T}$). Therefore, we explicitly employ geometric constraints in pose estimation formulation to improve the learning-based 3D object detection.


\begin{figure*}[ht!]
\centering
\includegraphics[width=0.85\linewidth]{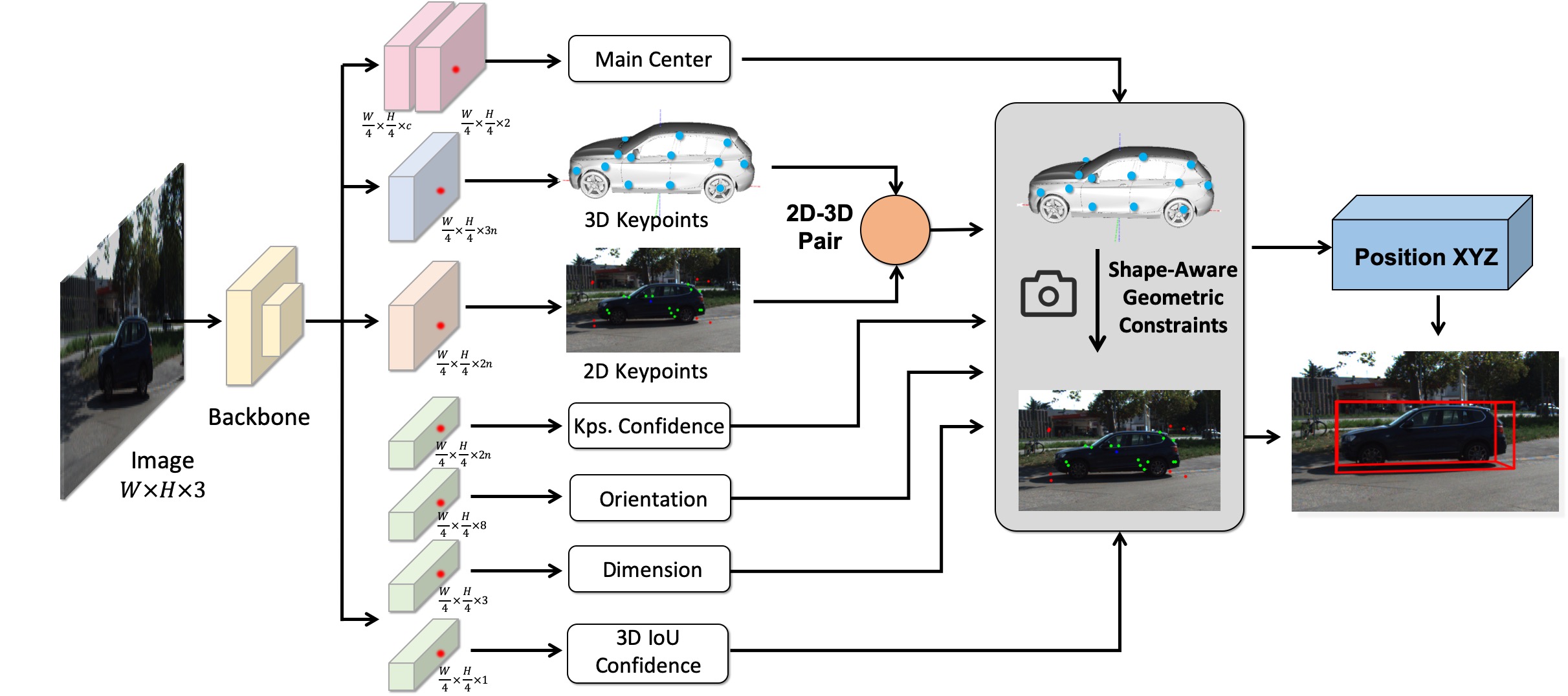}
\vspace{-1mm}
\caption{\textbf{Overview of the proposed keypoints-based 3D detection framework}. By passing the backbone network, 8 branch heads are followed for center point classification, center point offset, 2D keypoints, 3D coordinates, keypoints confidence, object orientation, dimension, and 3D detection score regression purpose. Finally, all the regressed information has been employed for recovering the object's 3D Bbox in the camera coordinate.}
\label{Fig:Network}
\end{figure*}

\section{Proposed Method}\label{sec:Approach}

In this section, we propose a general deep learning-based 3D object detection framework, which can employ the 2D/3D geometric constraints. To well explore the prior knowledge, CAD models are employed here. First, we pre-define several distinguished 3D keypoints on CAD models. Then, we propose to build the correlation between these 3D keypoints and their 2D projections on the image resorting to the deep learning network. Finally, the object pose can be easily solved with these geometrical constraints. 
More importantly, all the processes are implemented into the neural network, which can be trained in an end-to-end manner.

\subsection{Point-wise 2D-3D Constraints}\label{subsec:2d_3d_constraint}
Assuming a 3D point $\mathbf{P}_{o}^{i}$ $(x_{o}^{i}, y_{o}^{i}, z_{o}^{i})$ in object local coordinate, then its projection location $(u^{i}, v^{i})$ on the image plane can be obtained based on Eq. \ref{Eq:o2c} and Eq. \ref{Eq:c2imag} as
\begin{equation}
s \left [
    \begin{array}{c}
     u^{i}  \\ v^{i}  \\ 1 
   \end{array} 
   \right] 
   = \mathbf{K} 
   \left [ 
    \begin{array}{cc}
    \mathbf{R} & \mathbf{T}
    \end{array} 
   \right ]
    \left [ 
    \begin{array}{c}
    x_{o}^{i} \\
    y_{o}^{i} \\
    z_{o}^{i} \\
    1
    \end{array} 
\right ].
\label{Eq:o2image}
\end{equation}
In autonomous driving scenario, the road surface that the object lies on is almost flat locally, therefore the orientation parameters are reduced from three to one by keeping only the yaw angle $r_y$ around the Y-axis. Therefore the rotation matrix $\mathbf{R}$ becomes as \small  $ \setlength{\arraycolsep}{1.1pt} 
         \left [ \begin{array}{ccc}
        cos(r_y) & 0  & sin(r_y)  \\
        0 & 1 &  0  \\
        -sin(r_y) & 0 &  cos(r_y)  \\
    \end{array} \right ]$ \normalsize 
and Eq. \ref{Eq:o2image} can be simplified as 
\begin{small}
\begin{equation}
    \setlength{\arraycolsep}{1.0pt}
    \left [\begin{array}{c c c}
     -1  & 0  & \Tilde{u}^{i} \\
     0  & -1  & \Tilde{v}^{i} \\
   \end{array} \right] 
   \left [\begin{array}{c}
     T_x  \\ T_z  \\ T_z \
   \end{array} \right] 
   =  
   \left [ 
    \begin{array}{c}
    x_{o}^{i}cos(r_y) + z_{o}^{i}sin(r_y) + \\ \Tilde{u}^{i}[x_{o}^{i}sin(r_y) - z_{o}^{i}cos(r_y) ]
    \\
    y_{o}^{i} +  \Tilde{v}^{i}[x_{o}^{i}sin(r_y) - z_{o}^{i}cos(r_y)] \\
    \end{array} 
\right ],
\label{Eq:2d_3d_constraints}
\end{equation}
\end{small}
where  $\Tilde{u}^{i} = (u^{i} - c_x) / f_x $, $\Tilde{v}^{i} = (v^{i} - c_y) / f_y $ and $\mathbf{T} = [T_x, T_y, T_z]^{T}$. As described in Eq. \ref{Eq:2d_3d_constraints}, for each object point, two constraints are given. If $n$ points are provided, $n \times 2$ constraints can be obtained as 
\begin{equation}
    \mathbf{A} \mathbf{T} = \mathbf{B}, 
    \label{Eq:npoints_constraint}
\end{equation}
where 
\small 
\begin{equation*}
\mathbf{A} = 
  \left [
    \begin{array}{ccc}
        -1  & 0 & \Tilde{u}_{1} \\
        0 & -1 & \Tilde{v}_{1} \\
        & \vdots & \\
         -1  & 0 & \Tilde{u}_{n} \\
        0 & -1 & \Tilde{v}_{n} \\
    \end{array}
    \right ]_{2n \times 3 },
\end{equation*}
\normalsize
and 
\footnotesize 
\begin{equation*}
 \setlength{\arraycolsep}{1.0pt}
     \mathbf{B} =
    \left [
    \begin{array}{c}
        {x}_{o}^{1} \cos({r_{y}}) + {z}_{o}^{1} \sin({r_{y}})  +
        \Tilde{u}^{1}({x}_{o}^{1} \sin({r_{y}}) - {z}_{o}^{1} \cos({r_{y}}))
        \\
         y_{o}^{1}  +
        \Tilde{v}^{1}(x_{o}^{1} \sin({{r_y}}) - z_{o}^{1} \cos({{r_y}})) 
                \\
                        \vdots \\
        {x}_{o}^{n} \cos({r_{y}}) + {z}_{o}^{n} \sin({r_{y}})  +
        \Tilde{u}^{n}({x}_{o}^{n} \sin({r_{y}}) - {z}_{o}^{n} \cos({r_{y}}))
        \\
         y_{o}^{n}  +
        \Tilde{v}^{n}(x_{o}^{n} \sin({{r_y}}) - z_{o}^{n} \cos({{r_y}}))  
    \end{array}
    \right ]_{2n \times 1}.
\end{equation*}
\normalsize

However, in the real AD scenario, not all the keypoints can be seen from a certain camera viewpoint. For these keypoints which have been seriously occluded, the 2D/3D keypoints regression can not be well guaranteed. To well handle this kind of uncertainty, we propose to output an additional score to measure the confidence of each keypoint. And this score can be used as a weight during the pose calculation process. Specifically, for the $2n$ constrains in Eq. \ref{Eq:npoints_constraint}, an additional weights $c=\{c_1, c_2, ..., c_{2n}\}$ have been added to determine their importance during the pose calculation procedure. Therefore, Eq. \ref{Eq:npoints_constraint} can be reformulated as  
\begin{equation}
    \text{diag}(c)\mathbf{A} \mathbf{T} =  \text{diag}(c)\mathbf{B}
    \label{Eq:cofficient_npoints_constraint}
\end{equation}

In this linear system, $\mathbf{T}$ represents the object location in the camera coordinate system, which can be solved by providing 2D/3D correspondences and the rotation angle $r_y$. Here, the 3D keypoints are defined in the local object's coordinate varying in a relatively small range and the 2D keypoints are defined in the image domain. Both of them are easy for networks to learn. However, manual labeling of the ground truth for 2D and 3D keypoints is very costly and tedious. Therefore, we develop an auto-labeling pipeline by optimizing the 2D and 3D reprojection errors. The detailed annotation pipeline will be introduced in Section \ref{sec:autolabeling}.  





\subsection{Network}\label{subsec:Network}
An overview of our proposed framework is illustrated in Fig. \ref{Fig:Network}. Here, we follow one-stage-based 3D object detection framework such as CenterNet \cite{zhou2019objects} for its inference efficiency. Our proposed framework is backbone independent and here we employ DLA-34 \cite{yu2018deep} in our implementation. Given an image $\mathbf{I}$ with width $W$ and height $H$, the output feature map will be 4 times smaller than $\mathbf{I}$ after passing through the backbone network. To well utilize the geometric constraints, the following information is required to be learned from the deep neural network. 


\textbf{Object Center:} in anchor-free based object detection frameworks, the object center is essential information, which serves two functions: one is whether there is an object and the other is that if there exists an object, where is the center. Usually, these two functions are realized by a classification branch by distinguishing a pixel whether is an object center or not. The output of this branch will be $\frac{W}{4} \times \frac{W}{4} \times C $, where $C$ is the number of classes. 

Besides the classification, an additional ``offset'' regression branch is required to compensate for the quantization error during the down-sampling process. The output of this branch is $\frac{W}{4} \times \frac{H}{4} \times 2 $ to represent the offset in $x$ and $y$ direction respectively.         

\textbf{Object Dimension:} a separate branch is used to regress the object dimension $h$, $w$, $l$ with the output size of $\frac{W}{4} \times \frac{H}{4} \times 3$. Similar to other approaches, we don't regress the absolute object's size directly and regress a relative scale compared to the mean object size of each class. Details operation can be found in \cite{zhou2019objects}.

\textbf{2D Keypoints:} rather than directly detect these keypoints from the image, we regress $n$ ordered 2D offset coordinates for each object center. The benefit is that the number and order of keypoints for each object can be well guaranteed. In addition, the regression for the offset is easier by removing the object center. The output size of this branch is $\frac{W}{4} \times \frac{H}{4} \times 2n$.

\textbf{3D Keypoints:} similar to the 2D keypoints, we regress the 3d keypoints in the local object coordinate. In addition, all 3D keypoints values are normalized by object dimension $(l, w, h)$ in $x$, $y$, $z$-direction respectively. By using this format, the 3D keypoints values are in a relatively small range, which will benefit the whole regression process. The output size of this branch is $\frac{W}{4} \times \frac{H}{4} \times 3n$.

\textbf{Object Orientation:} similarly, we regress local orientation angle with respect to the ray through the perspective point of 3D center following Multi-Bin based method \cite{mousavian20173d}. Here, 8 bins are used with the output size of $\frac{W}{4} \times \frac{W}{4} \times 8$.

\textbf{Keypoints Confidence Scores:} for each keypoint, a couple of additional confidence scores have been regressed for measuring its contribution in the linear system for solving the object pose. For $2n$ constrains in Eq . \ref{Eq:npoints_constraint}, a feature map with size of $\frac{W}{4} \times \frac{W}{4} \times 2n$ will be outputted. 

\textbf{3D IoU Confidence Score:} rather using the classification score directly as the object detection confidence, we add one branch to regress the 3D IoU score in purpose. This score is supervised by the IoU between estimated Bbox and ground truth Bbox. Finally, the product of this score and the output classification score is assigned as the final 3D detection confidence score.

\subsection{Loss Function}\label{subsec:Losses}
The overall loss contains the following items: a center point classification loss $l_m$ and center point offset regression loss $l_{off}$, a 2D keypoints regression loss $l_{2D}$, a 3D keypoints points regression loss $l_{3D}$, an orientation multi-bin loss $l_r$, a dimension regression loss $l_D$, a 3D IoU confidence loss $l_c$ and a 3D bounding box IoU loss $l_{IoU}$. Specifically, the multi-task loss is defined as
\begin{equation}
\begin{split}
    \mathbf{L} = w_m l_m + w_{off} l_{off} + w_{2D} l_{2D} + w_{3D} l_{3D} + \\ w_r l_r + w_D l_D + w_c l_c + w_{IoU} l_{IoU}
\end{split}
\end{equation}
where $l_m$ is the focal loss as used in \cite{zhou2019objects}, $l_{2D}$ is a depth-guided $l-1$ loss as used in \cite{2009.00764}, $l_D$ and $l_{3D}$ are L1 loss with respect to the ground truth. Orientation loss $l_r$ is the Multi-Bin loss. 3D IoU confidence $l_{c}$ is a binary cross-entropy loss supervised by the IoU between the predicted 3D BBox and ground truth. $l_{IoU}$ is the IoU loss between the predicted 3D BBox and ground truth \cite{zhou2019iou}.

\begin{figure*}[ht!]
\centering
\includegraphics[width=0.8\linewidth]{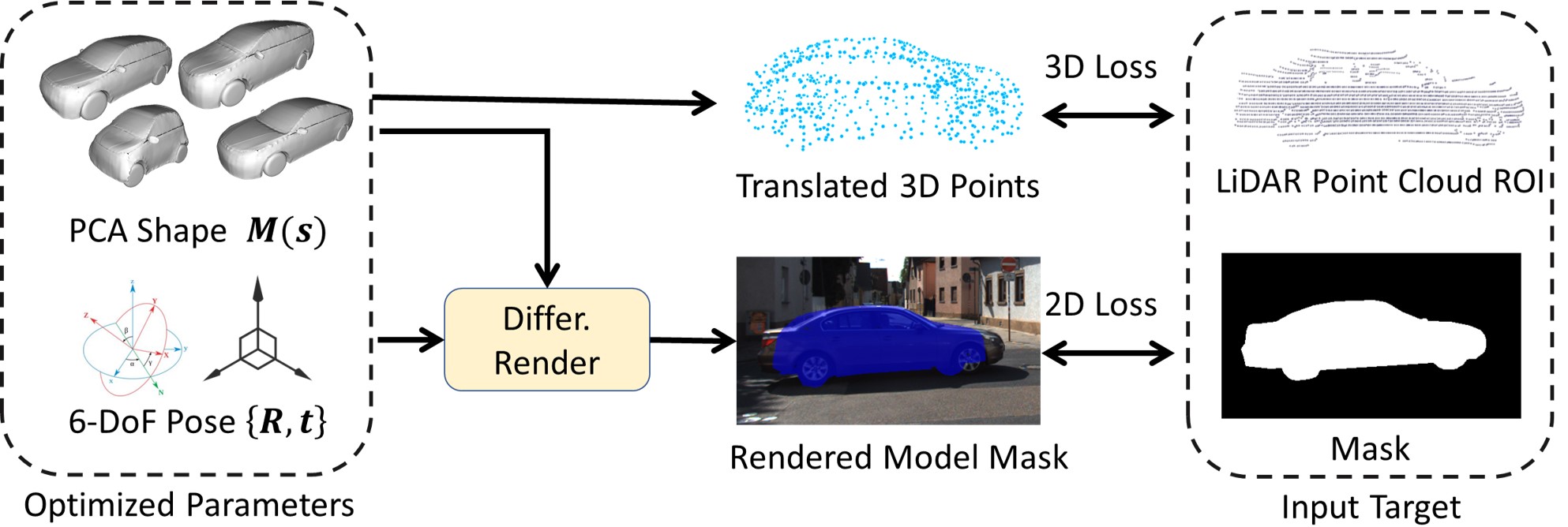}
\caption{\textbf{The pipeline of the proposed 3D shape-aware auto-labeling framework}. By providing different kinds of vehicle CAD models, a mean shape template and $r$ principal basic can be obtained. By giving a 3D object sample in the KITTI dataset, the optimal principal components with coefficient $s$ and 6-DoF pose can be iteratively optimized by minimizing the 2D and 3D losses which are defined on the scanned sparse point cloud and the instance mask. }
\label{Fig:Auto_shape_Overview}
\end{figure*}


\section{3D Shape Auto-Labeling}\label{sec:autolabeling}
In this section, we will introduce how to automatically fit the 3D shape to the visual observations and then automatically generate ground-truth annotations of 2D keypoints and 3D locations in the local object coordinate for training the network. The main process is illustrated in \ref{Fig:Auto_shape_Overview}. Different from existing methods which use only a few CAD models for 3D labeling (\eg, 11 in \cite{zakharov2020autolabeling} and 16 in \cite{menze2015object}), we adopt a 3D deformable vehicle template~\cite{song2019apollocar3d, lu2020permo} that can represent arbitrary vehicle shape by adjusting the parameters. Therefore, the 3D shape labeling process can be formulated as an optimization problem that aims at computing the optimal parameter combination to fit the visual observations (\ie 2D instance mask, 3D bounding box, and 3D LiDAR points).



\subsection{Deformable Vehicle Template}
\label{sec:Deformable_vehicle_Representation}
In the real-world traffic scenarios, there are many different vehicle types (\eg, coupe, hatchback, notchback, SUV, MPV, \etc) and their geometric shapes vary significantly. To perform 3D shape fitting, a straight-forward solution is to build a 3D shape dataset and the fitting process can be regarded as model retrieval. However, dataset construction is labor-intensive, inefficient, and costly. Instead, we use a deformable 3D model for vehicle representation~\cite{lu2020permo}. Specifically, this template is composed of a set of PCA (Principal Components Analysis) basis $r$. Any new 3D vehicle $\mathcal{M}(s)$ can be represented as a mean shape model $\mathcal{M}_0$ plus a linear combination of $r$ principal components with coefficient $s=[s_1, s_2, ..., s_r]$ as
\begin{equation}
 \mathcal{M}(s) = \mathcal{M}_0 + \sum_{k=1}^{r}s_k\delta_kp_k,
	\label{equ:pcamodel}
\end{equation}
where $p_k$ and $\delta_k$ are the principal component direction and corresponding standard deviation and $s_k$ is the coefficient of the $k_{th}$ principal component. Based on the vehicle template, we can automatically fit an optimal 3D shape to the visual observations (details in Subsec. \ref{subsec:shape_optimization}).

\begin{figure}[ht!]
\centering
\includegraphics[width=0.98\linewidth]{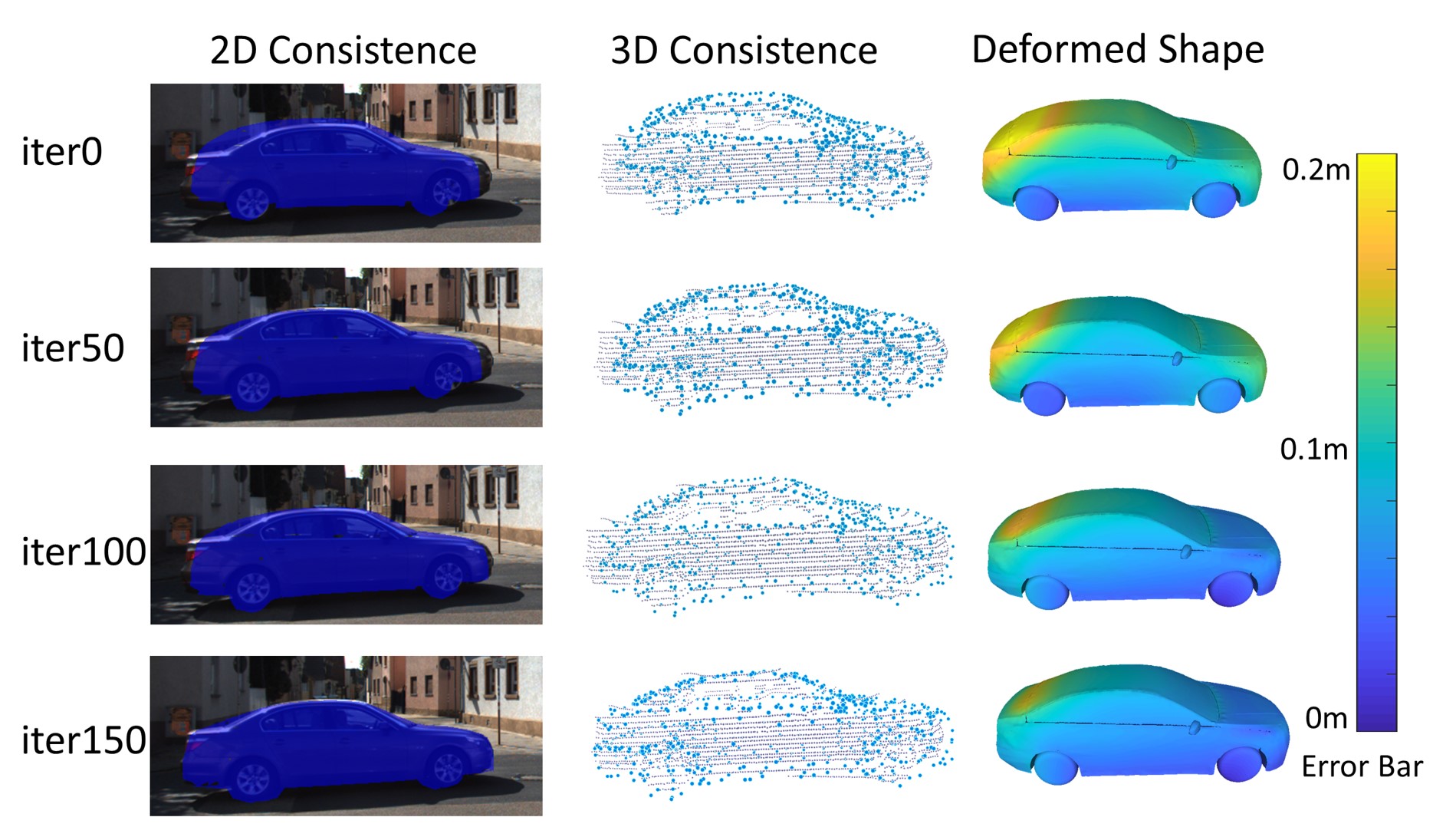}
\caption{\textbf{Illustration of the 3D shape optimization process} from step 0 to step 200. From the top to bottom, the rendered mask gradually covers the target mask and 3D model vertices align with the point cloud gradually and the 3D deformed shape changes from a mean shape to a `notchback'.}
\label{Fig:label_process}
\vspace{-3mm}
\end{figure}

\subsection{3D Shape Optimization}\label{subsec:shape_optimization}

For each vehicle, our goal is to assign a proper 3D shape to fit the visual observations, including the 2D instance mask, the 3D bounding box, and the 3D LiDAR points. Specifically, the annotations of 2D instance mask $I_{ins}$ and the 3D bounding box $B_{box}$ are provided by the KINS dataset \cite{qi2019amodal} and the KITTI dataset \cite{geiger2012we}, respectively. The annotation of 3D LiDAR points for each vehicle is much more complex. According to the labeled 3D bounding boxes, we first segment out the individual 3D points from the entire raw point cloud. Then we remove the ground points using the ground-plane estimation method (\ie RANSAC-based plane fitting). Finally, we obtain the ``clean'' 3D points for each vehicle, which is represented as $\mathbf{p} = \{p_0, ..., p_k\}$.

\begin{figure}[ht!]
\centering
\includegraphics[width=0.9\linewidth]{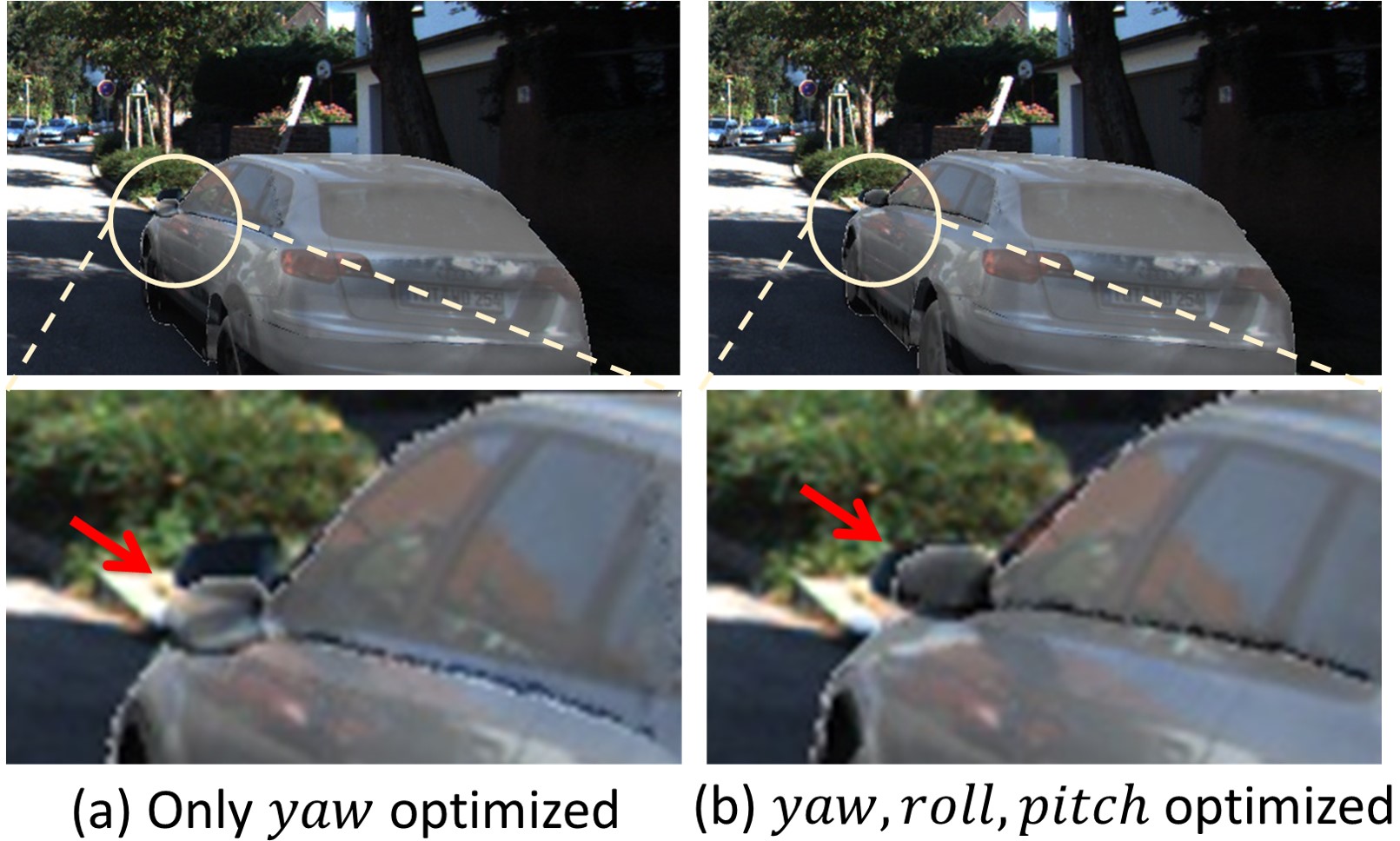}
\caption{\textbf{An example of CAD model-fitting results} with one angle ($yaw$) only or three angles ($yaw$, $roll$, and $pitch$). When the road is not flat, serious misalignment will happen if only $yaw$ angle is applied for optimization.
}
\label{Fig:angle_optimization}
\end{figure}

The 3D shape annotation is to compute the best PCA coefficient $\mathbf{\widehat{s}}$ and the object 6-DoF pose ($\widehat{\mathbf{R}}, \widehat{\mathbf{t}}$). Existing 3D object detection benchmarks (\eg, KITTI, Waymo) only label the yaw angle because they assume vehicles are on the road plane. However, we experimentally (an example has been given in Fig. \ref{Fig:angle_optimization}) find that the other two angles (\ie pitch and roll) can significantly improve the 3D shape annotation results. Therefore, the loss function is formulated as
\begin{equation}
\begin{split}
    \widehat{\mathbf{s}}, \widehat{\mathbf{R}}, \widehat{\mathbf{t}} = \mathop{\arg\min}_{\mathbf{s}, \mathbf{R}, \mathbf{t}} \{ \alpha \mathbf{L}_{2D} \left ( Pr \left (\mathcal{M} \left ({s} \right ), {\mathbf{R}}, {\mathbf{t}} \right ), I_{ins} \right ) \\ + \beta \mathbf{L}_{3D} \left ( v \left ( {s} \right ), \mathbf{R}, \mathbf{t}, \mathbf{p} \right ) \}, 
\end{split}
\label{equ:autoshape_optimzatgion}
\end{equation}
which consists of 3D points loss $\mathbf{L}_{3D}$ and 2D instance loss $\mathbf{L}_{2D}$. $\alpha$, $\beta$ are two hype-parameters to balance these two constraints. 
The operation $Pr\left ( \cdot  \right )$ is a differentiable rendering function to produce the binary mask $\tilde{I}_{ins}$ of $\mathcal{M} \left ({s} \right )$ with translation $\{ {\mathbf{R}}, {\mathbf{t}} \}$. Specifically, the $\mathbf{L}_{2D}$ is defined as the sum distance over each pixel $(i,j)$ in image $I$

\begin{equation}
    \mathbf{L}_{2D} = \sum_{(i,j)\in I}\left \| \tilde{I}_{ins}(i,j) - I_{ins}(i,j) \right \|.
     \label{equ:loss2d}
\end{equation}

For each 3D point $p_i \in \mathbf{p}$, we find its nearest neighbor $v_i$ in $v(\tilde{s})$ and compute the distance between them. $\mathbf{L}_{3D}$ is defined as the sum distance over all correspondences pairs 
\begin{equation}
    \label{equ:loss3d}
    \mathbf{L}_{3D} = \sum_{ p_i \in \mathbf{p} }\left \| p_i - v(\tilde{s})_i \right \|.
\end{equation}
The function of Eq.~\ref{equ:autoshape_optimzatgion} can be optimized by the gradient descent strategy. The vehicle's center position and orientation are used for $\mathbf{t}$ and yaw angle initialization, while the pitch, roll angles, and PCA coefficients are initialized as zeros. Then we forward the pipeline and compute $\mathbf{L}_{2D}$ and $\mathbf{L}_{3D}$ and finally back-propagate gradients to update $s$, $\mathbf{R}$, $t$. In Fig.~\ref{Fig:label_process}, we depict the intermediate result during the optimization process.


\section{Experimental Results}\label{sec:exp}

We implement the approach and evaluate it on the public KITTI \cite{geiger2012we} 3D object detection benchmark. 
\subsection{Dataset and Implementation Details}
\textbf{Dataset:} the KITTI dataset is collected from the real traffic environment from the Europe streets. The whole dataset has been divided into training and test two subsets, which consist of $7,481$ and $7,518$ frames, respectively. Since the ground truth for the test set is not available, we divide the training data into a \textit{train} set and a \textit{val} set as in \cite{zhou2018voxelnet}, and obtain $3,712$ data samples for training and $3,769$ data samples for validation to refine our model. On the KITTI benchmark, the objects have been categorized into ``Easy'', ``Moderate'', and ``Hard'' based on their height in the image and occlusion ratio, \etc.

\textbf{Evaluation Metric:} we focus on the evaluation on ``Car'' category because it has been considered most in the previous approaches. For evaluation, the average precision (AP) with Intersection over Union (IoU) is used as the metric for evaluation.
Our AutoShape approach is compared with existing methods on the \textit{test} set using $AP_{R_{40}}$ by training our model on the whole $7,481$ images. We evaluate on the \textit{val} set for ablation by training our model on the \textit{train} set using $AP_{R_{40}}$.

\begin{table}[t!]
	\centering
	\resizebox{0.485\textwidth}{!}
	{%
		\begin{tabular}{r| l c c c c c c c}
			\hline
			\multicolumn{1}{c}{\multirow{2}{*}{Methods}} & \multicolumn{1}{c}{\multirow{2}{*}{Modality}} & \multicolumn{3}{c}{\textbf{$\textbf{AP}_\text{3D}$70 (\%)}} & \multicolumn{3}{c}{\textbf{$\textbf{AP}_\text{BEV}$70 (\%)}} & \multicolumn{1}{c}{\multirow{2}{*}{Time (s)}} \\
			\multicolumn{1}{c}{} & \multicolumn{1}{c}{} & \multicolumn{1}{l}{Moderate} & \multicolumn{1}{l}{Easy} & \multicolumn{1}{l}{Hard} & \multicolumn{1}{l}{Moderate} & \multicolumn{1}{l}{Easy} & \multicolumn{1}{l}{Hard} & \multicolumn{1}{c}{}\\ \hline
			M3D-RPN \cite{brazil2019m3d} &Mono & 9.71 & 14.76 & 7.42 &   13.67 & 21.02 & 10.23 & 0.16 \\
			SMOKE \cite{liu2020smoke} &Mono  & 9.76 & 14.03 &  7.84 & 14.49 & 20.83  &  12.75 & 0.03\\
			MonoPair \cite{chen2020monopair} &Mono  & 9.99 & 13.04 &  8.65 &14.83 & 19.28  &  12.89  & 0.06\\
			RTM3D \cite{li2020rtm3d} & Mono  & 10.34 & 14.41 & 8.77 & 14.20 & 19.17  &  11.99 & 0.05\\
			AM3D \cite{ma2019accurate} &$\text{Mono}^{\ast}$ & 10.74 & 16.50 &  9.52 & 17.32 & 25.03  &  14.91 & 0.4 \\
            PatchNet \cite{ma2020rethinking} & $\text{Mono}^{\ast}$   & 11.12 & 15.68 & 10.17 & 16.86 &  22.97 &  14.97 & 0.4\\ 
			RefinedMPL \cite{vianney2019refinedmpl} & $\text{Mono}^{\ast}$   & 11.14& 18.09 & 8.94 & 17.60 &  28.08 &  13.95 & 0.15\\ 

			KM3D \cite{2009.00764} & Mono    & 11.45 & 16.73 & 9.92 & 16.20 & 23.44  &  14.47 & 0.03\\
			D4LCN \cite{ding2020learning} &$\text{Mono}^{\ast}$   & 11.72 & 16.65 &  9.51 & 16.02 & 22.51  &  12.55 & 0.20\\

			IAFA \cite{zhou2020iafa} &Mono   & 12.01 & 17.81 & 10.61  &  17.88 & 25.88 & 15.35    & 0.03\\ 
			YOLOMono3D\cite{liu2021yolostereo3d} & Mono  & 12.06 & 18.28 & 8.42 & 17.15 & 26.79  &  12.56  & 0.05 \\
            Monodle\cite{ma2021delving} & Mono  & 12.26 & 17.23 & 10.29 & 18.89 & 24.79  &  16.00 & 0.04 \\
            MonoRUn\cite{chen2021monorun} & Mono  & 12.30 & 19.65 & 10.58 & 17.34 & 27.94  &  15.24 & 0.07 \\
            GrooMeD-NMS\cite{kumar2021groomed} & Mono  & 12.32 & 18.10 & 9.65 & 18.27 & 26.19  &  14.05 & 0.12 \\
            DDMP-3D\cite{wang2021depth} & Mono  & 12.78 & 19.71 & 9.80 & 17.89 & 28.08  &  13.44 & 0.18 \\

            Ground-Aware\cite{liu2021ground} & Mono  & 13.25 & 21.65 & 9.91 & 17.98 & 29.81  &  13.08 & 0.05 \\
            CaDDN\cite{reading2021categorical} & $\text{Mono}^{\ast}$ & 13.41 & 19.17 & 11.46 & 18.91 & 27.94  &  \B{17.19} & 0.63 \\
            MonoEF\cite{zhou2021monocular} & Mono & 13.87 & 21.29 & \B{11.71} & 19.70 & 29.03  &  \R{17.26} & 0.03 \\
            MonoFlex\cite{zhang2021objects} & Mono & \B{13.89} & 19.94 & \R{12.07} & \B{19.75} & 28.23  &  16.89 & 0.03 \\

            \hline

		    Baseline Method\cite{2009.00764} & Mono  & 11.45 & 16.73 & 9.92 & 16.20 & 23.44  &  14.47 & 0.03\\
		    \textbf{AutoShape-16kps} & Mono$\dagger$ & 13.72 & \B{21.75} & 10.96  &  19.00 & \B{30.43} & 15.57 & 0.04 \\
		    \textbf{AutoShape-48kps} & Mono$\dagger$ & \R{14.17} & \R{22.47} & 11.36  &  \R{20.08} & \R{30.66} & 15.59 & 0.05 \\
		    Improvements & - & \textbf{+2.72} & \textbf{+5.74} & +\textbf{1.44}  &  \textbf{+3.88} & \textbf{+7.22} & \textbf{+1.12}  & - \\
			\hline  
		\end{tabular}
	}
	\caption{ \textbf{Comparison with other public methods on the KITTI testing server for 3D ``Car'' detection}. For the ``direct'' methods, we represent the `` Modality'' with ``Mono'' only. We use $\ast$ to indicate that the ``depth'' has been used by these methods during training and inference procedure. $\dagger$ indicates that 'CAD models' have been used in data labeling stage. For easy understanding, we have highlighted the top numbers in \R{red} for each column and the second best is shown in \B{blue}. 
	}
	\label{tab:test_kitti}
	\vspace{-3mm}
\end{table}

\textbf {Implementation Details:} 
We implement our auto-labeling approach (Sec.~\ref{sec:autolabeling}) using differentiable renderer \cite{kato2018neural, liu2019softras, liu2020general} 
which is optimized by using the Adam optimizer with a learning rate of 0.002. To speed the optimization and save memory, we downsample the PCA model to $666$ vertices and $998$ faces. We set $\alpha$ and $\beta$ to 1.0 and 5.0, respectively.
Our shape-aware 3D detection network uses DLA-34~\cite{yu2018deep} as backbone. We pad the image size to $1280 \times 384$. 3D IoU confidence loss weight $w_c$ and 3D IoU loss weight $w_{iou}$ are increased from 0 to 1 with exponential RAMP-UP strategy \cite{laine2016temporal}. We use Adam optimizer with a base learning rate of 0.0001 for 200 epochs and reduce by $10 \times$ at 100 and 160 epochs. We project the ground truth to corresponding right image and use random scaling (between 0.6 to 1.4), random shifting in the image range, and color jittering for data augmentation. The network is trained on 2 NVIDIA Tesla V100 (16G) GPU cards and the batch size is set to 16. For the KITTI \textit{test} set evaluation, we sample 16/48 keypoints from the 3D shape, 8 corner points, and 1 center to train  network.

\subsection{Data Auto-Labeling Evaluation} 

Our approach can automatically generate the 2D keypoints and their corresponding 3D locations in the local object coordinate which are employed as the supervision signal during the training process. To verify the quality of the labeling results, the 2d instance segmentation mean AP and 3D bounding box mean AP is used here for verification. Specifically, the 2D instance segmentation IoU is calculated using the projected mask by 3D models and the ground truth mask (from KINS \cite{qi2019amodal}). We obtain the labeled 3D bounding box using the dimension of the 3D model with the optimized 6-DoF pose, which is compared to the ground-truth 3D bounding box provided by \cite{geiger2012we} with the mean IoU score. Tab.~\ref{tab:autoshape_acc} shows the detailed comparison results. The proposed method can achieve 0.86 for 2D mean AP and 0.76 for 3D mean AP, which justifies the effectiveness of our auto-labeling approach.

\subsection{Evaluation for 3D Object Detection}
The evaluation of the proposed approach with other SOTA methods for 3D detection detection on KITTI~\cite{geiger2012we} \textit{test} set are given in Tab.~\ref{tab:test_kitti}. 
From the table, we can obviously find that the proposed method with 48 keypoints achieves \R{4} first places in 6 tasks 
with the $AP|_{R_{40}}$ metric. We also report our method with 16 keypoints, which has faster inference time and keeps promising accuracy. 
In addition, most of the existing methods such as \cite{ma2019accurate, ma2020rethinking, vianney2019refinedmpl, ding2020learning, reading2021categorical}, need to estimate the depth map, resulting in a heavy computation burden in inference. In contrast, our method obtains the depth information by 3D shape-aware geometric constraints, which is more accurate with faster running speed. We achieve 25 FPS with an NVIDIA V100 GPU card with 16 keypoints configuration. Compared with baseline geometric constraint methods~\cite{2009.00764, li2020rtm3d} using 8 corners and 1 center point as keypoints for training, our method with 48 keypoints utilizes more shape-aware keypoints to construct stronger geometric constraints, getting \textbf{+5.74\%}, \textbf{+2.72\%}, \textbf{1.44\%}, \textbf{+7.22\%}, \textbf{+3.88\%}, \textbf{+1.12\%} improvements for $AP_{3D}$ and $AP_{BEV}$ on ``Easy'', ``Moderate'', and ``Hard'' categories.

\subsection{Qualitative Results}
\begin{figure}[ht!]
\centering
\includegraphics[width=0.475\textwidth]{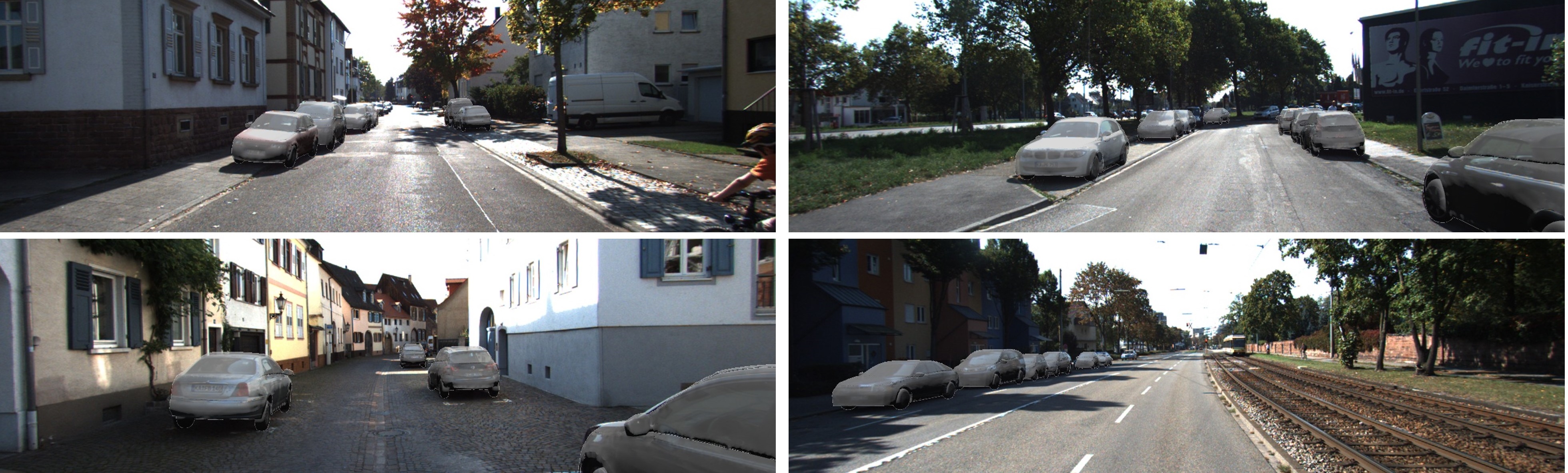}
\caption{Qualitative results of our 3D shape auto-labeling.}
\label{Fig:autolabel_res}
\vspace{-5mm}
\end{figure}

\begin{figure}[ht!]
\centering
\includegraphics[width=0.475\textwidth]{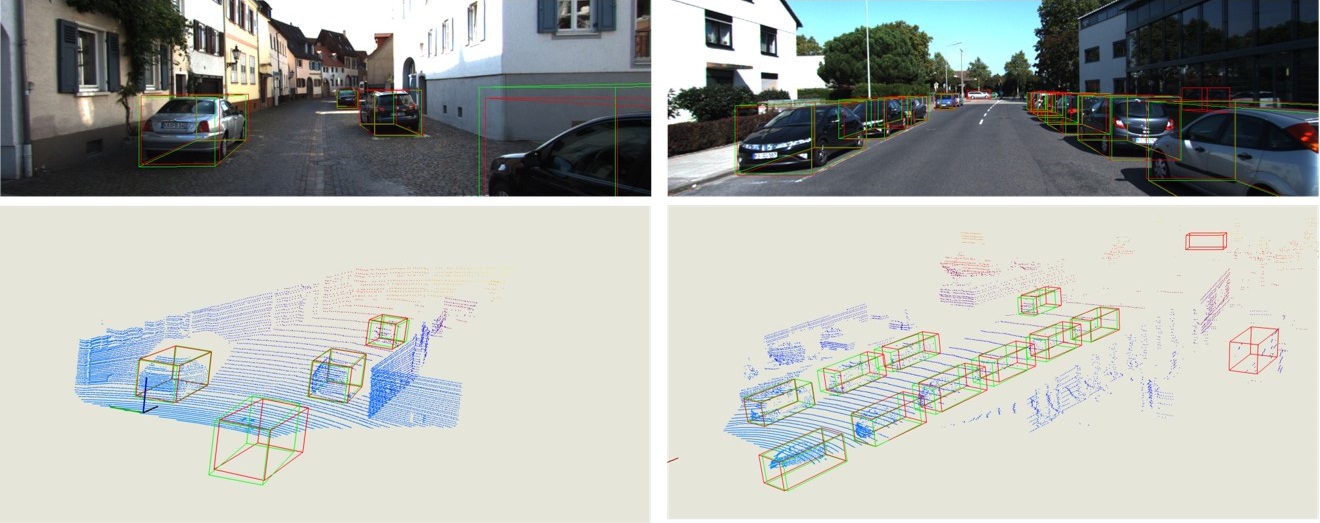}
\caption{Qualitative 3d detection results on KITTI \textit{validation} set. Red boxes represent our predictions, and green boxes come from ground truth. LiDAR signals are only used for visualization.}
\label{Fig:3ddetctionr_res}
\vspace{-3mm}
\end{figure}

Qualitative results of 3D shape auto-labeling are shown in Fig \ref{Fig:autolabel_res}. Each vehicle in the image is overlaid with a rendered 3D model optimized by our method. We can see the consistency of our labeled shape and the real object. We also visualize some representative results of our shape-aware model in Fig \ref{Fig:3ddetctionr_res}. Our model can predict object location accurately even for distant and truncated objects.

\subsection{Ablation Studies}

\begin{figure}[ht!]
\centering
\includegraphics[width=0.43\textwidth]{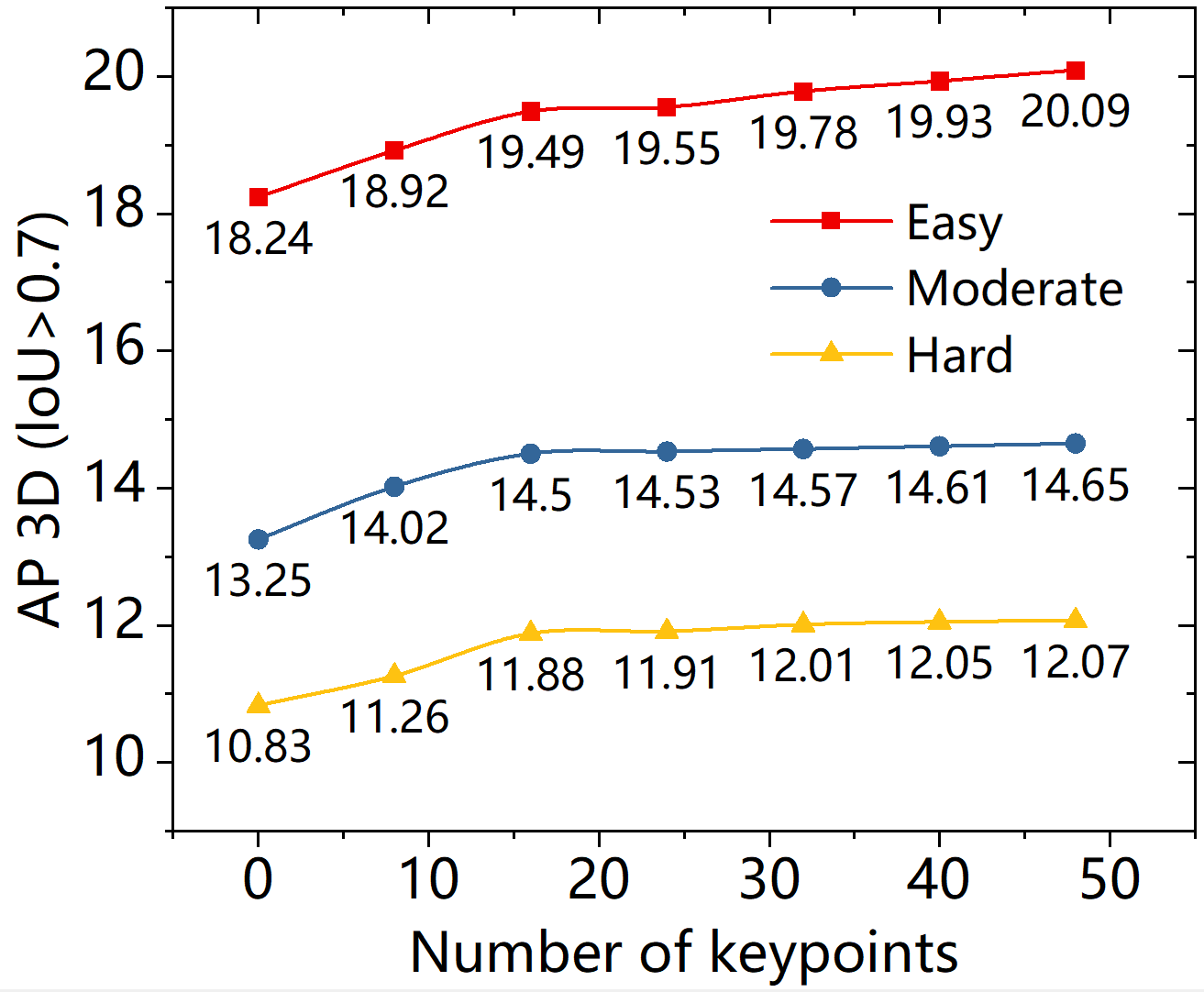}
\caption{3D object detection performance with different number of keypoints on KITTI \textit{val} set using $AP|_{R_{40}}$.}
\label{Fig:Number_of_Points}
\vspace{-3mm}
\end{figure}

\textbf{The Number of Keypoints:} our shape-aware 3D detection network benefits from the geometric constraint of 2D-3D keypoints from the 3D shape. To better understand the effect of different numbers of the keypoints, we set it from 0 to 48 with an interval of 8. Note that the 8 corners and 1 center point are always maintained in this experiment and we vary the extra keypoints. As shown in Fig.~\ref{Fig:Number_of_Points}, from 0 to 16, the network performance is significantly improved. 
From 16 to 48, however, we observe that the network performance is not sensitive to the number of the keypoints. The main reason is that more dense 2D shape points can be overlapped in the $\frac{W}{4} \times \frac{H}{4}$ heatmap during the regression process.  Furthermore, with more keypoints, the network consumes more GPU memory for storage and computation, resulting in longer training and inference time. In practice, we set the number of extra keypoints to 16, which is a good compromise of accuracy and efficiency.

\textbf{2D/3D Loss for Auto-Labeling:} our auto-labeling approach (Sec.~\ref{sec:autolabeling}) can generate precise posed 3D shape for each 2D vehicle instance. The key technique is simultaneously optimizing the 2D/3D constraints (loss) for better matching. Here, we conduct an ablation study to justify the effectiveness of the 2D/3D loss. We first only use 3D point loss $\mathbf{L}_{3D}$ in the objective function. Then we only use the 2D mask loss $\mathbf{L}_{2D}$ for optimization. Finally, we take both 2D/3D loss into computation. Tab.~\ref{tab:autoshape_acc} shows that using both 2D/3D loss get the best performance in the auto-labeling process. We further observe that the impact of 2D mask loss $\mathbf{L}_{2D}$ is more important than 3D point loss. By using both $L_{2D}$ and $L_{3D}$, the labeling accuracy is improved to 86.35 and 76.92, resulting in better 3D detection performance. This correlation indicates that 3D detection performance can be significantly improved by using high-quality labeling data of 3D shapes.

\begin{table}[ht!]
	\centering
	\resizebox{0.4\textwidth}{!}
	{%
    \begin{tabular}{cc|cc|ccc}
        \hline
        \multirow{2}{*}{$\mathbf{L}_{2D}$} & \multirow{2}{*}{$\mathbf{L}_{3D}$} & \multicolumn{2}{c|}{Label Acc.}                & \multicolumn{3}{c}{Car 3D Det.}                             \\ \cline{3-7} 
                                  &                            & \multicolumn{1}{c|}{$I_{2D}$} & $I_{3D}$ & \multicolumn{1}{c|}{Easy} & \multicolumn{1}{c|}{Mod.} & Hard \\ \hline
                                & \checkmark                         & 0.61                           & 0.71      & 15.49                         & 11.35                         & 9.34    \\
        \checkmark                        &                          & 0.82                           & 0.74     & 18.36                        & 13.88                       & 11.23   \\
        \checkmark                        & \checkmark                         & \textbf{0.86}                           & \textbf{0.76}      & \textbf{20.09}                        & \textbf{14.65}                        & \textbf{12.07}   \\ \hline
        \end{tabular}
        
    }
\caption{Shape Autolabeling Ablation Experiments on KITTI \textit{val} set using $AP|_{R_{40}}$.}
\label{tab:autoshape_acc}
\end{table}

\vspace{-0.5cm}
\section{Conclusion}
In this paper, we present a framework for real-time monocular 3D object detection by explicitly employing shape-aware geometric constraints between 3D keypoints and their 2D projections on images. Both the 3D keypoints and 2D project points are learned from deep neural networks. We further design an automatic annotation pipeline for labeling object 3D shape, which can automatically generate the shape-aware 2D/3D keypoints correspondences for each object. Experimental results show our approach can achieve state-of-the-art detection accuracy with real-time performance. Our approach is general for other types of vehicles, and in the future, we are interested in validating the performances of our approach on other objects.

{\small
\bibliographystyle{ieee_fullname}
\bibliography{egbib}
}
\section{Supplemental Material}\label{sec:supplements}

\subsection{Ablation Study for Keypoints Confidence Regression}
The 2D/3D keypoints regression is a critical component in the proposed framework, however, inaccurate regression of these keypoints is inevitable in the real AD scenario due to many reasons e.g., viewpoint change, occlusion, and labeling noise, etc. Especially, these prediction outliers will greatly affect the results of the linear system described in Eq. \ref{Eq:npoints_constraint}. In order to handle this problem, we propose to
predict a confidence score for each keypoint and employ it as a weight for determining its contribution to the linear system. To verify the effectiveness of the prediction confidence, we set a series of ablations studies on the ``Car'' category.

\begin{figure}[ht!]
\centering
\includegraphics[width=0.9\linewidth]{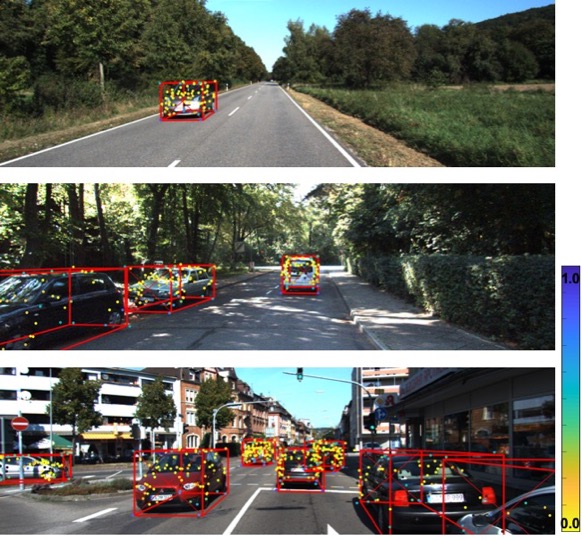}
\caption{Visualization of keypoints confidences. Here, the blue represents score ``1'' and yellow represents score ``0'' and the color changing from blue to yellow represents the confidence score decreasing from ``1'' to ``0''. This figure is better to view in color print.}
\label{Fig:vis_kp}
\end{figure}

We give the results in Tab. \ref{tab:keypoints_confidence}. From this table, we can see that the 3D object detection performance can be significantly improved by integrating the regressed key points confidences. More importantly, this improvement is independent of the number of keypoints. In addition, for further understanding the actual meaning of this predicted confidence, we have visualized them in Fig. \ref{Fig:vis_kp}. Interestingly, we find that these keypoints with high confidences usually come from the ground point (the intersection point between the tire and the ground) and these distinguished shape border points. These points will give more contribution to the object pose estimation.




\begin{table}[ht!]
\centering
\resizebox{0.4\textwidth}{!}
	{%
\begin{tabular}{c|c|l|l|l}
\hline
\multicolumn{1}{l|}{\multirow{2}{*}{Num. Kps.}} & \multirow{2}{*}{Kps. Confi.} & \multicolumn{3}{c}{Car 3D Det.} \\ \cline{3-5} 
\multicolumn{1}{l|}{}                          &                            & Easy      & Mod.      & Hard     \\ \hline
\multirow{2}{*}{16}                            & \multicolumn{1}{l|}{}      & 16.49     & 12.31     & 10.54    \\ \cline{2-5} 
                                               & \checkmark                        & \textbf{19.59}     & \textbf{14.50}     & \textbf{11.88}    \\ \hline
\multirow{2}{*}{48}                            & \multicolumn{1}{l|}{}      & 16.85     & 12.39     & 10.04    \\ \cline{2-5} 
                                               & \checkmark                        & \textbf{20.09}     & \textbf{14.65}     & \textbf{12.07}    \\ \hline
\end{tabular}
}
\caption{Keypoints confidence ablation experiments on KITTI \textit{val} set using $AP|_{R_{40}}$ metric.}
\label{tab:keypoints_confidence}
\end{table}

\subsection{Multi-classes Detection}

Currently, the designed Autoshape model can't generate the keypoints annotation for ``Pedestrian'' and ``Cyclist'' due to the lack of CAD models. Here, we simply transform the 3D keypoints from the mean ``Car'' template to the ``Pedestrian'' and ``Cyclist'' by normalize them first and re-scale them to the bounding box's size of other categories. By generating these keypoints, then the object's pose can easily solve as the ``Car'' category. We evaluate multi-class 3d detection on the KITTI \textit{test} sever and the performances are shown in Tab. \ref{tab:multi-class detetion}. From this table, we can find that the proposed framework performs relatively well even though the keypoints annotation is not very accurate of ``Pedestrian'' and ``Cyclist''. Interestingly, we find that the cyclist gives much better results than the ``Pedestrian'' and this is because the ``Cyclist'' can be considered as a rigid object to some extent. On the contrary, the ``Pedestrian'' is a non-rigid object and the location of these keypoints varies a lot with different object pose.  

\begin{table}[ht!]
\centering
\resizebox{0.485\textwidth}{!}
	{%
\begin{tabular}{c|c|c|c|c|c|c}
\hline
\multirow{3}{*}{Methods} & \multicolumn{6}{c}{3D Det.}                             \\ \cline{2-7} 
                         & \multicolumn{3}{c|}{Pedestrian} & \multicolumn{3}{c}{Cyclist} \\ \cline{2-7} 
                         & Easy      & Mod.      & Hard     & Easy     & Mod.     & Hard    \\ \hline
M3D-RPN\cite{brazil2019m3d}                  & 4.92      & 3.48     & 2.94     & 0.94     & 0.65    & 0.47    \\
MonoPair\cite{chen2020monopair}                   & \textbf{10.02}      & \textbf{6.68}     & \textbf{5.53}     & 3.79     & 2.12    & 1.83    \\
MonoFlex\cite{zhang2021objects}                 & 9.43     & 6.31     & 5.26     & 4.17     & 2.35    & 2.04    \\ \hline
Ours                     & 5.46      & 3.74     & 3.03     & \textbf{5.99 }    & \textbf{3.06}    & \textbf{ 2.70}    \\ \hline
\end{tabular}
}
\caption{Quantitative results for ``Pedestrian`` and ``Cyclist`` on KITTI \textit{test} set with $AP|_{R_{40}}$ metric.}
\label{tab:multi-class detetion}
\end{table}

\end{document}